\documentclass[aaai]{article}
\usepackage{aaai}
\usepackage{times}
\usepackage{helvet}
\usepackage{courier}
\usepackage{graphicx}
\usepackage{xcolor} 
\usepackage[normalem]{ulem} 
\begin{document}
%
\title{Investigating the Effects of Robot Proficiency Self-Assessment \\ on Trust and Performance}
\author{Nicholas Conlon$^{1}$, Daniel Szafir$^{2}$, and Nisar Ahmed$^{3}$\\
$^{1,2}$Department of Computer Science, $^{3}$Smead Aerospace Engineering Sciences\\
$^{1,3}$University of Colorado Boulder, $^{2}$University of North Carolina at Chapel Hill\\
nicholas.conlon@colorado.edu, daniel.szafir@cs.unc.edu, nisar.ahmed@colorado.edu\\
}
\newcommand{\nickComm}[1]{{\color{red}{ \textbf{NJC:} #1}}}
\newcommand{\brettComm}[1]{{\color{blue}{ \textbf{BWI:} #1}}}
\newcommand{\nisarComm}[1]{{\color{magenta}{ \textbf{NRA:} #1}}}

\maketitle
\thispagestyle{empty}
\pagestyle{empty}

\begin{abstract}
Human-robot teams will soon be expected to accomplish complex tasks in high-risk and uncertain environments. Here, the human may not necessarily be a robotics expert, but will need to establish a baseline understanding of the robot’s abilities in order to appropriately utilize and rely on the robot. This willingness to rely, also known as trust, is based partly on the human’s belief in the robot’s proficiency at a given task. If trust is too high, the human may push the robot beyond its capabilities. If trust is too low, the human may not utilize it when they otherwise could have, wasting precious resources. In this work, we develop and execute an online human-subjects study to investigate how robot proficiency self-assessment reports based on Factorized Machine Self-Confidence affect operator trust and task performance in a grid world navigation task. Additionally we present and analyze a metric for trust level assessment, which measures the allocation of control between an operator and robot when the human teammate is free to switch between teleportation and autonomous control. Our results show that an \textit{a priori} robot self-assessment report aligns operator trust with robot proficiency, and leads to performance improvements and small increases in self-reported trust.
\end{abstract}

%
\section{INTRODUCTION} \label{INTRO}
Appropriate use of autonomy is key for the success of human-robot teams in high risk and uncertain scenarios. Examples include battlefield surveillance and targeting \cite{naval_urban_uav,history_of_uavs}, forest fire detection \cite{uav_fire_detection,uav_fire_detection_monitoring,dist_wildfire}, natural disaster assessment \cite{disaster_imaging,disaster_management}, and space exploration \cite{lunar_mining,exploration_robots_space}. Human-robot teams in these domains are expected to operate in unconstrained, dynamic, and uncertain environments, and may not be co-located with their team or support elements.

As motivation, consider a navigation task assigned to a human-robot team. The robot, an Unmanned Ground Vehicle (UGV) is located on site while its human teammate (the operator) is at an offsite command center. On the ground, the robot has a clear picture of local obstacles and dangers. However due to bandwidth constraints it may not be able to relay all of this information to the operator. Here the operator has two options for UGV control: (1) rely on the UGV to autonomously execute the task (or some part of it), or (2) switch to a high bandwidth teleportation control and execute the task (or part of it) themselves. Losing the robot can be costly if not mission ending, so the operator must balance the information they receive with their trust in the UGV executing the task. For example if the operator has a high certainty that there are a large number of craters nearby or debris which may cause sensor degradation, they may choose to teleoperate that section and then return control to the UGV when the terrain becomes more forgiving. On the other hand, if the operator has high confidence in the UGV's proficiency, they may let it traverse the entire environment autonomously.

Trust can be thought of as \emph{a psychological state in which an agent willingly and securely becomes vulnerable, or depends on, a trustee, having taken into consideration the characteristics (e.g., benevolence, integrity, competence) of the trustee.} \cite{israelsen2019dave}. In our motivating scenario, the appropriate use of the robot was rooted in trust. Having too much trust in the system could cause the operator to overestimate the robots proficiency, and push it beyond its capabilities, while not enough trust can lead to the autonomy sitting idle which could increase human workload and waste precious resources \cite{misuse_disuse}. In order for an operator to develop the appropriate level of trust towards a robotic system, they need an understanding of the robot's proficiency at a given task.

Early work in proficiency assessment developed the notion of self-confidence (or self-trust) as the agents belief in its ability to meet mission goals \cite{Hutchins2015}. The authors develop a sensing-optimization/verification-action (SOVA) model to capture the agents self-confidence as well as a Trust Annunciator Panel to display it to an operator. Building off their work, Factorized Machine Self-Confidence (FaMSeC) \cite{israelsen2019ConfPaper,aitken2016Thesis} developed a computational framework for proficiency self-assessment in autonomous agents. FaMSeC allows an agent to assess its task self-confidence relative to a users expectations by analyzing a set of related factors impacting proficiency (task interpretation, model quality, model validity, expected outcome, past experiences). Promising preliminary results show an improvement to both trust and performance when applied to a human supervisory task  \cite{israelsenThesis}. In this study we look directly at how reporting an \textit{a priori} FaMSeC Outcome Assessment to an operator impacts both task performance and trust in a simple navigation task.

\section{EXPERIMENTAL DESIGN}\label{METHODS}
This study was approved by the University of Colorado Boulder Internal Review Board and conducted online through Amazon Mechanical Turk.

\subsection{Navigation Task}
A participant was teamed with a simulated robot and tasked with navigating the robot through a grid world from a starting location to a goal location. There were two autonomy levels which the participant could toggle between at any time during the task: \textit{Manual Control} let the participant drive the robot manually via the a user interface. \textit{Automatic Control} let the robot autonomously drive itself towards the goal. 

The environment contained obstacles, debris, and craters. Obstacles could not be driven through. Debris temporarily degraded the robots navigation accuracy and caused the robot to inadvertently move to an adjacent free space in the grid if driven over while in \emph{Automatic Control} mode. Craters captured the robot and result in immediate task failure if driven over in either autonomy level. A task \textit{configuration} was a specific placement of obstacles, debris, and craters.

\begin{figure}[htbp]
\centering
\includegraphics[width=8cm]{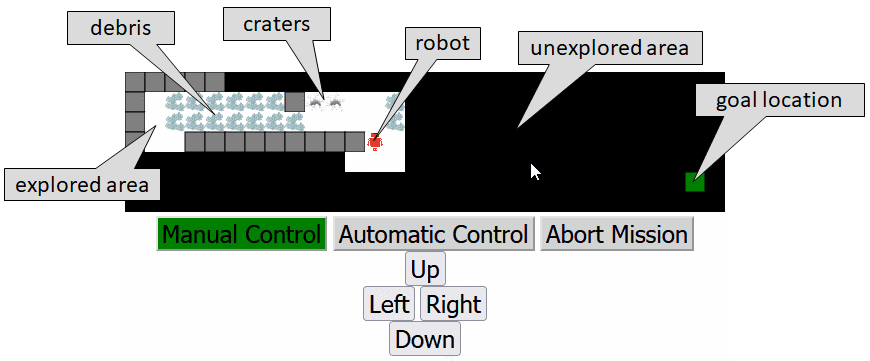}
\caption{Interface displayed to the participant. The map (top) showed obstacles (obstacles, debris, craters), 
the robot location (e.g. red robot), the goal (finish), exported and unexplored areas. Below were buttons
to choose autonomy level (\textit{Manual Control} or \textit{Automatic Control}) or abort the task.
Below were buttons for the participant to pilot the robot if they are in \textit{Manual Control} mode.}
\label{fig:INTERFACE}
\end{figure}

The participant interacted with the simulated robot via the interface shown in Fig. \ref{fig:INTERFACE}. The top half of the display contained a map which always showed the robots location and the goal location. The map display also showed any obstacles present within the robot's immediate area (sensor FOV), and any previously explored areas. All unexplored areas of the map were dark. Below the map were buttons for \textit{Manual Control} and \textit{Automatic Control} which allowed the operator to change between the two mutually exclusive autonomy modes. Next to the control buttons was an \textit{Abort Mission} button which allowed the participant to end the task at anytime if they believe it necessary. At the bottom of the interface were buttons for the participant to control the robot (up, down, left, right) if in \textit{Manual Control} mode.

\subsection{Formulation} The task was formulated as a Markov Decision Process to support \textit{Automatic Control} of the robot.

\textbf{State space $(S)$:} We represented the state $s \in (x,y)$ as the $(x, y)$ position of the robot, where $x \in [0,30]$ $y \in [0,7]$ was the size of the grid world.

\textbf{Action space $(A)$:} There were four actions $a \in [$Up, Down, Left, Right$]$, which moved the robot one space in the respective direction in the grid world.

\textbf{Rewards $(R)$:} The robot received a small negative reward of $-1$ for executing an action, a large positive reward of $+100$ for driving onto the goal location, and a large negative reward of $-100$ for driving into a crater (and failing the task). This simple reward structure encouraged the robot to move directly to the goal while avoiding obstacles.

\textbf{Transition function $(T)$:} The transition function moved the robot from state $s$ to state $s'$ when executing action $a$. If the state $s'$ contained an obstacle, then the robot stayed in state $s$ with probability $1$. If the state $s'$ contained debris, the robot instead moved to an open location adjacent to $s'$ (above, below, left, or right) with equal probability. The robot moved to $s'$ with probability $1$ if $s'$ did not contain an obstacle or debris. 

We compute the value iteration policy $\pi_{robot,i}$ for task configuration $i$  offline. 

\subsection{Robot Self-Assessment}

For each task configuration the robot runs $100$ simulations with $\pi_{robot, i}$ to generate a distribution of expected rewards. The reward distribution $R$ was then used to compute a reward-based FaMSeC Outcome Assessment (OA) statement, which indicated the robots belief in its ability to achieve rewards greater than a user specified reward threshold $R_{min}$. This was accomplished by computing the upper partial moment-lower partial moment $UPM/LPM$ ratio with respect to $R_{min}$ as in Eqn. \ref{eqn:upm_lpm}.
\begin{equation}
    \frac{\textit{UPM}}{\textit{LPM}}|_{R_{min}} = \frac{\int_{R_{min}}^\infty r p(r) dr}{\int_{-\infty}^{R_{min}} r p(r) dr}
    \label{eqn:upm_lpm}
\end{equation}
In operational settings a user would choose $R_{min}$. However for this study we fixed $R_{min}$ at $0$. That is, we considered tasks in which robot attained greater then or equal to $0$ rewards desirable, and tasks in which the robot attained less than $0$ rewards to be undesirable. The $UPM/LPM$ is in the range $[-\inf, +\inf]$, so we next transformed to the range $[-1,+1]$ by Eqn. \ref{eqn:oa}.
 \begin{equation}
     \textit{OA} = \frac{2}{1+e^{\left(\frac{\textit{UPM}}{\textit{LPM}}|_{R_{min}}\right)}} - 1
     \label{eqn:oa}
 \end{equation}
The resultant metric \textit{OA} was then mapped to a semantic label according to the \textit{OA} statement column of Table \ref{tab:oa_info} and reported to the user in sentence form:  
\begin{quote}
   \emph{"the $<$color$>$ robot has $<$statement$>$ confidence in navigating to the goal"}. 
\end{quote}
For example, \emph{"The green robot has good confidence in navigating to the goal"} may be presented to the participant if they were teamed with a green robot and that robot had an outcome assessment of \textit{good} for this task configuration. If the robot self-assessment was present, it appeared just above the map in the user interface. All possible statements a robot could report along with their descriptions can be seen in Table \ref{tab:oa_info}. 

\begin{table}[]
    \centering
    \begin{tabular}{|p{0.23\linewidth} ||p{0.17\linewidth}|| p{0.43\linewidth}|}
    \hline
    \textbf{\textit{OA} Range} & \textbf{Statement} & \textbf{Description} \\
    \hline
    \hline
    [-1,-0.75) & very bad & The robot believes it will \textit{greatly fall short} of user expectations while in Automatic Control mode\\
    \hline
    [-0.75, -0.25) & bad & The robot believes it will \textit{fall short} of user expectations while in Automatic Control mode\\
    \hline
    [-0.25, 0.25] & fair & The robot believes there is an \textit{equal chance of exceeding or falling short} of user expectations while in Automatic Control mode\\
    \hline
    (0.25, 0.75] & good & The robot believes it will \textit{exceed} user expectations while in Automatic Control mode\\
    \hline
    (0.75, 1] & very good & The robot believes it will \textit{greatly exceed} user expectations while in Automatic Control mode \\
    \hline
    \end{tabular}
    \caption{Table of Outcome Assessment range, statement semantic label, and statement description for each self-assessment statement that could be reported by the robot.}
    \label{tab:oa_info}
\end{table}

\subsection{Factors and Levels}
Our study was a mixed-design with $2 \times 2 \times 1=4$ conditions.

\begin{enumerate}
    \item Factor 1 (between-subjects) was robot reporting accuracy. This factor had two levels: (1) \textbf{informed reports}, where the robot reported an Outcome Assessment statement computed using the FaMSeC Outcome Assessment formula, and (2) \textbf{random reports}, where the robot reported a statement chosen uniformly at random from all possible statements. Random reports simulated a robot having a poor self-assessment ability.
    
    \item Factor 2 (between-subjects) was robot driving performance. This factor had two levels and was only experienced by the participant while in \textit{Automatic Control} mode: (1) \textbf{high performance}, where the robot chose each action following the policy $\pi_{robot}$, and (2) \textbf{random performance}, where the robot chose an action following $\pi_{robot}$ with probability $p=0.5$, and uniformly random over all actions otherwise. 
   
    \item Factor 3 (within-subjects) was a base case where the robot displayed no reports to the participant. We referred to this as \textbf{absent report}.
\end{enumerate}

We randomly assigned reporting accuracy level and robot performance level at the beginning of the study. The participant was not told of which level they had been assigned. Other than a training round with a high performing robot, all other robots the participant encountered during the study preformed and reported according to the assigned levels. Each participant encountered a set of tasks to navigate to the goal. For each set of tasks the robot either reported its \textit{a priori} self-assessment or reported no information. 

\subsection{Procedure}
Each Amazon Mechanical Turk participant was given a link which took them to our web page. The web page contained an overview of the study, IRB information, and the option to participate. If the participant chose to participate they were then taken to a small suitability survey, and if passed, taken to a page containing a deeper overview of the study. This overview contained descriptions of the scenario and outcome assessment statements, instructions, and training on the interface. They were then give a small quiz on the information presented. Once complete they were given a single round of training where they could practice changing autonomy level and controlling the robot through the interface.

Each participant then encountered groups of navigation tasks. For each group of tasks the robot either reported its \textit{a priori} self-assessment or reported no information. The order of the groups was randomized as well as the task configuration (obstacles locations) within the group. After completing a group of tasks the participant was given a self-reported trust survey.

We developed a bonus pay structure to encourage the participant to rely on the autonomy when they thought it appropriate and to complete the task quickly. Participants started each task with $5$ points and the value was reduce to a minimum possible score of $0$. We payed participants a base rate consistent with federal minimum wage plus a bonus based on how many bonus points they had accumulated. The rules for bonus points are outlined in table \ref{BONUS}.

\begin{table}[h]
    \begin{center}
    \begin{tabular}{|p{0.4\linewidth} || p{0.4\linewidth}|}
    \hline
    \textbf{Event} & \textbf{Points Awarded} \\
    \hline
    \hline
    Fail task & $-5$ points\\
    \hline
    Abort task & $-3$ points \\
    \hline
    Drive in manual mode & $-0.1$ points per action \\
    \hline
    \end{tabular}
    \end{center}
    \caption{Bonus structure per participant. Each task started with 5 points and was reduced according to these rules.}
    \label{BONUS}
\end{table}

After completing all tasks, the participant was given a short demographics survey and provided information regarding payment. This completed the study.

\subsection{Measurements}
For each participant we captured two levels of data: In-mission data captured outcome, time, paths, and actions along paths for each task, while self-reported trust responses were captured after the participant completed a group of tasks (for example a group of tasks where the robot did not report a self-assessment). Each measure along with its associated description can be found in Table \ref{measures_table}.

\begin{table}[h]
    \begin{center}
    \begin{tabular}{|p{0.17\linewidth}||p{0.70\linewidth}|}
    \hline
    \textbf{Measure} & \textbf{Description} \\
    \hline
    \hline
    Outcome & The categorical task outcome: \emph{success}, \emph{failure}, or \emph{abort}.\\
    \hline
    Total time & Task completion time in seconds.\\
    \hline
    Participant actions &  $(x,y,a)$, where $(x, y)$ was the location where the participant executed action $a$ while in Manual Control mode.\\
    \hline
    Robot actions & A tuple $(x,y, a)$ where $(x, y)$ was the location where the robot executed action $a$ while in Automatic Control mode.\\
    \hline
    Self-reported trust &  The participants response to the MDMT Performance questionnaire after a set of $4$ tasks within the same condition.\\
    \hline
    \end{tabular}
    \end{center}
    \caption{Name and description for each measurement captured during the study.}
    \label{measures_table}
\end{table}

\subsection{Research Questions} 
First we were interested in task Outcome. Outcome was the categorical outcome of a task: (1) \emph{success} occurred when the team navigated the robot to the goal, (2) \emph{abort} occurred when the participant ended the task by choice, and (3) \emph{failure} occurred when the robot fell into a crater (either while the participant was driving or the autonomy was driving). We were interested in two main questions:

\begin{enumerate}
    \item Does outcome improve when the team is assigned a robot of \textbf{high performance}?
    \item Does outcome improve when the robot presents an \textbf{informed report}?
\end{enumerate}

Second we were interested in self-reported trust. Self-Reported Trust was the participants perception of their trust in the robot. We used the Multidimensional Measure of Trust V1 (\textit{Performance} dimension) questionnaire \cite{mdmt_ullman}. We had two main questions:

\begin{enumerate}
\setcounter{enumi}{2}
    \item Does trust increase when the team is assigned a robot of \textbf{high performance}?
    \item Does trust increase when the robot presents an \textbf{informed report}?
\end{enumerate}

Third we estimated Objective Trust through observations of trusting behavior during the task. If robot self-assessment reports aligned trust within the team, then we expected to observe increased participant actions when the robot reported lower confidence, and increased autonomy actions when the robot reported higher confidence. We developed a metric partly based on control allocation defined by \cite{team_performance} which we called \textit{control proportion}. This was the ratio of the difference over the sum of actions executed by the robot and the participant and can be seen in Eqn. \ref{eqn:CONT_PROP}.

\begin{equation}
    control\_proportion = \frac{a_{robot} - a_{participant}}{a_{robot}+a_{participant}}
    \label{eqn:CONT_PROP}
\end{equation}

This metric ranged from $[-1,1]$ and gave us insight into the average level of autonomy used during task execution. A $control\_proportion$ tending towards $-1$ indicated more participant actions than autonomy action, and conversely a $control\_proportion$ tending towards $+1$ indicated more autonomy actions than participant actions. Given this metric we can investigate the following question:

\begin{enumerate}
    \setcounter{enumi}{4}
    \item Does the presence of robot self-assessment reports cue the participant to change autonomy to an appropriate level?
\end{enumerate}

\subsection{Demographics}
We recruited $155$ total participants on Amazon Mechanical Turk. $74$ participants identified as male and $81$ identified as female. Ages ranged from $23$ to $77$ $(M=45.42, SD=10.81)$. Education ranged from some college classwork to graduate level degrees.

\section{RESULTS} \label{RESULTS}

In this section we discuss a subset of our experimental results pertaining to each of our research questions. All significance testing used an $\alpha = 0.05$. 

\begin{figure}[htbp]
    \centering
    \includegraphics[width=0.4\textwidth]{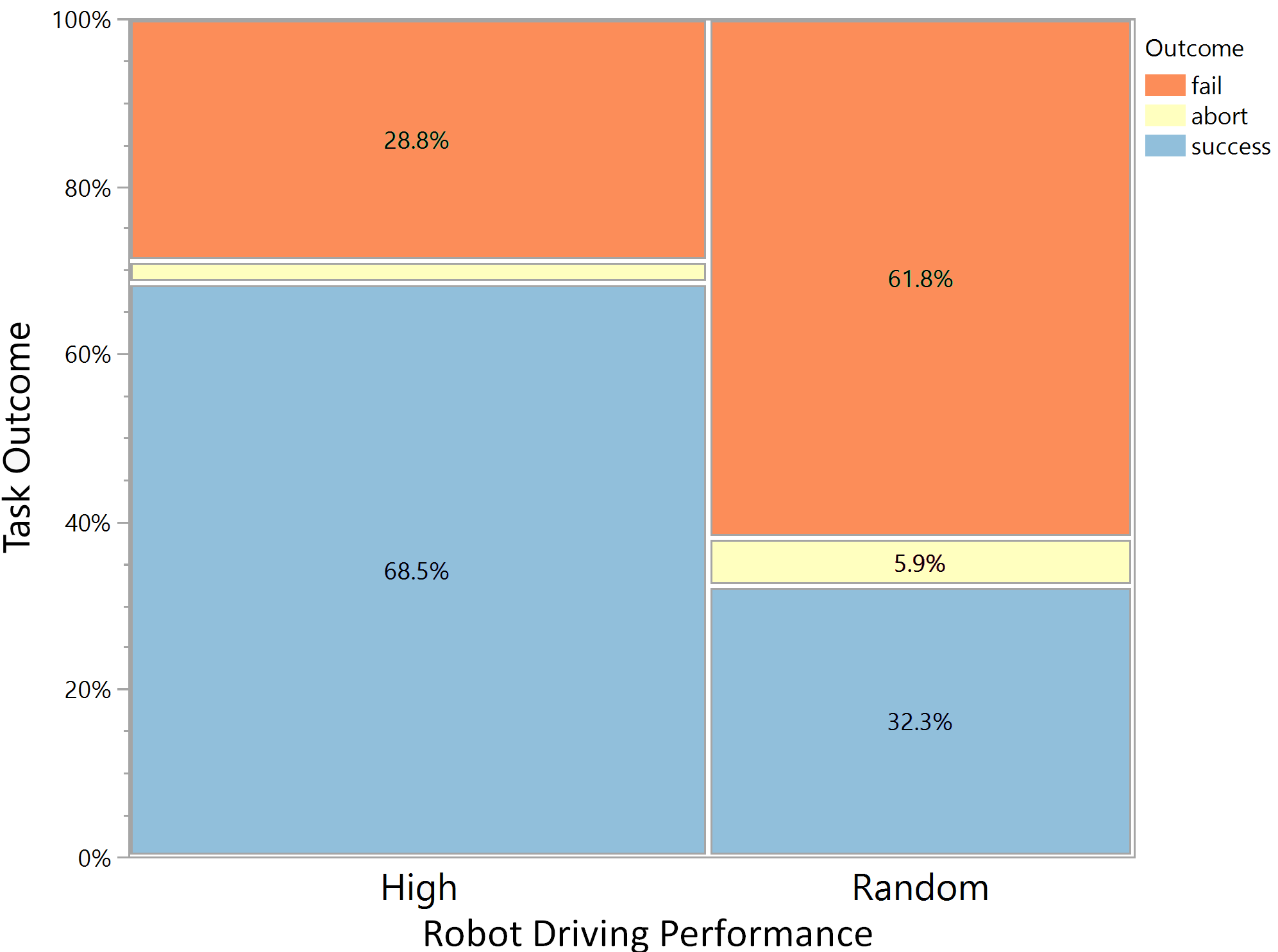}
    \caption{Stacked column plot of task outcome for high and random robot performance showing a statistically significant increase in task success when the participant was paired with a high performing robot.}
    \label{fig:OUTCOME_VS_PERFORMANCE}
\end{figure}

\emph{RQ1: Does outcome improve when the team is assigned a robot of \textbf{high performance}?} Fig. \ref{fig:OUTCOME_VS_PERFORMANCE} shows a stacked column plot of outcome for high and random robot performance while in Automatic Control mode. A contingency analysis showed that the team had significantly fewer failures when the agent operated at a high level of performance $(\chi^2(2, 1237)=164.02, p<.0001)$, with a moderate effect size according to Cramer's V (V=0.35). This indicated that outcomes did improve (teams were less likely to fail) when the team was assigned a high performing robot. 

\begin{figure}[htbp]
    \centering
    \includegraphics[width=0.4\textwidth]{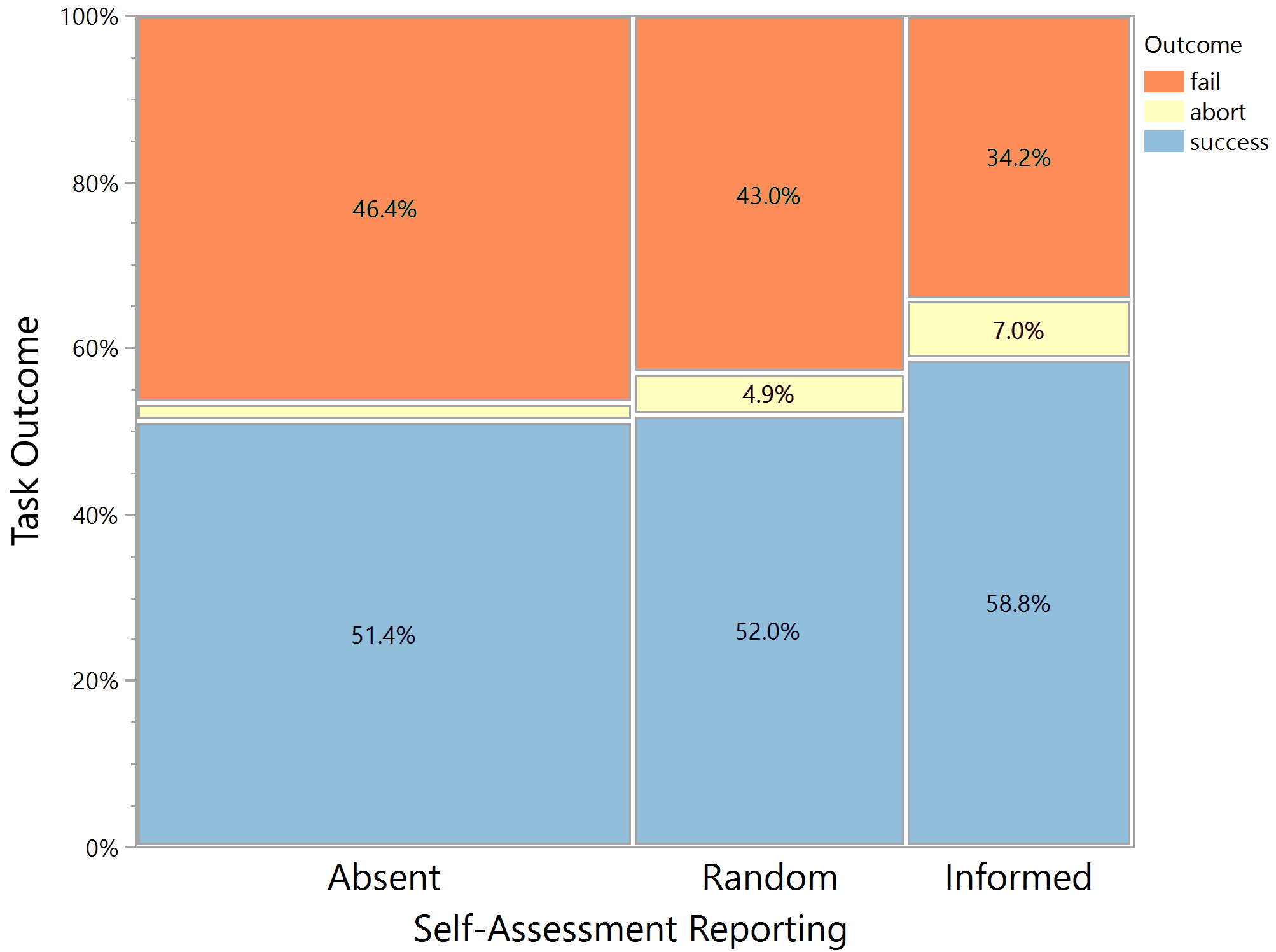}
    \caption{Stacked column plot of task outcome when self-assessments were absent, random, or informed showing a significant decrease in task failure between the informed and absent condition.}
    \label{fig:OUTCOME_VS_ACCURATE}
\end{figure}

\emph{RQ2: Does outcome improve when the robot presents an \textbf{informed report}?}  Fig. \ref{fig:OUTCOME_VS_ACCURATE} shows a stacked column plot of outcome with absent, random, and informed self-assessments. A contingency analysis between absent and informed showed that the team had significantly fewer failures when informed robot self-assessments were reported $(\chi^2(2, 904)= 19.813, p<.0001)$, with a moderate effect size according to Cramer's V (V=0.22). Additionally we observed that aborts increased when a robot self-assessment was present which may be due to the participant having an indication of task difficulty implicit in the robot's assessment. For reference the figure includes random self-assessment reports, however we did not include it in our analysis for this specific question.

\begin{figure}[htbp]
    \centering
    \includegraphics[width=0.4\textwidth]{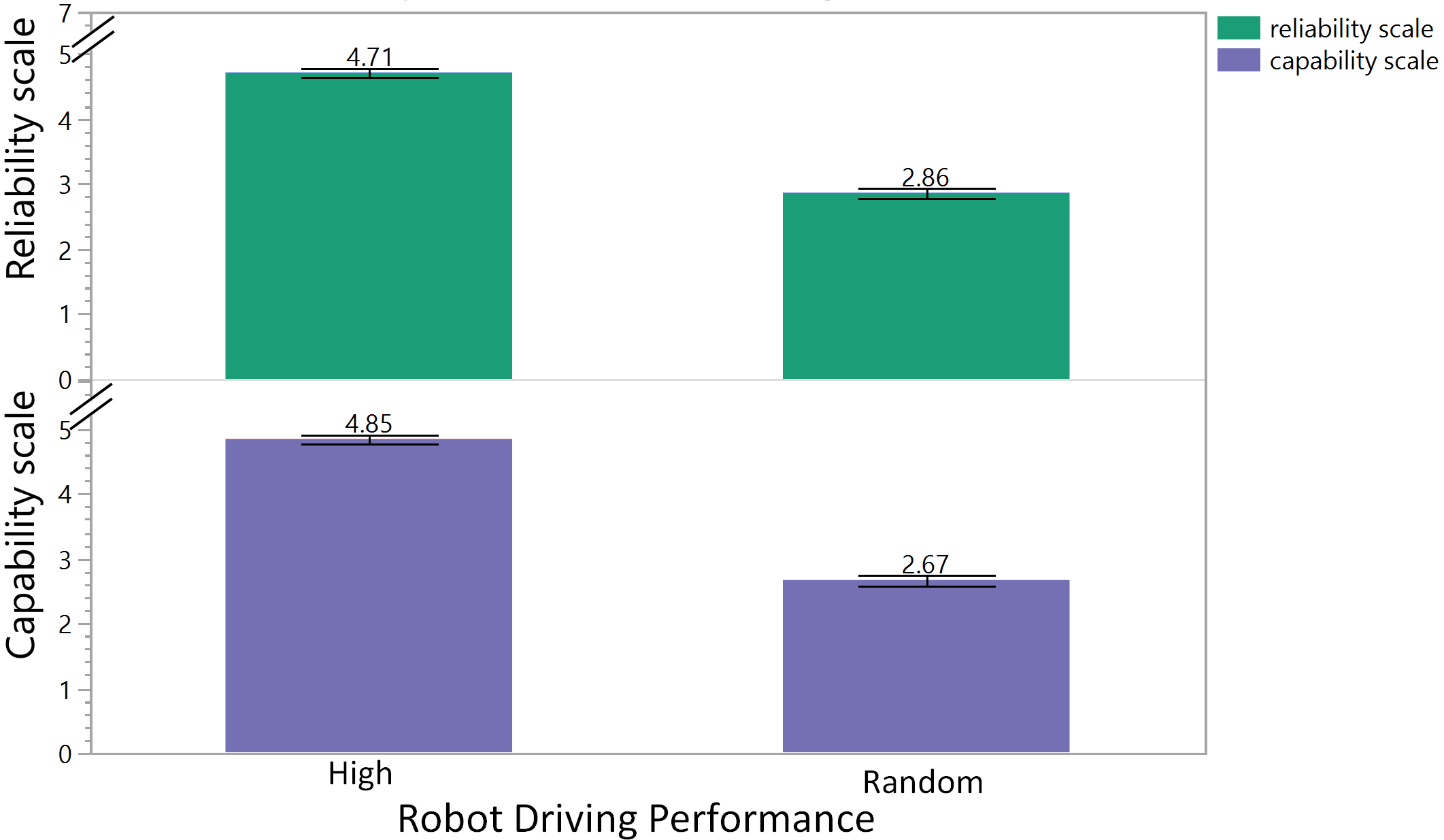}
    \caption{Bar plot of self-reported MDMT Reliability and Capability trust ratings for high and random robot performance showing a statistically significant increase in participant trust when the participant was paired with a high performing robot.}
    \label{fig:TRUST_VS_PERFORMANCE}
\end{figure}

\emph{RQ3: Does trust increase when the team is assigned a robot of \textbf{high performance}?} Fig. \ref{fig:TRUST_VS_PERFORMANCE} shows a bar plot of self-reported MDMT Reliability and Capability trust ratings for high and random robot performance while in Automatic Control mode. We conducted a one-tailed independent t-test to test whether the agent performance level affected participants self-reported MDMT capability and reliability trust scores and found that participants had significantly higher reliability trust when working with a high performing agent, $t(1149.7)=17.5$, $p<.0001$, with a large effect size according to Cohen's d (d=1.0). We also found significantly higher capability trust when working with a high performing agent, $t(1093.34)=19.73$, $p<.0001$, with a large effect size according to Cohen's d (d=1.13). This indicates that the presence of a higher performing robot could lead to substantial increases in trust.

\begin{figure}[htbp]
    \centering
    \includegraphics[width=0.4\textwidth]{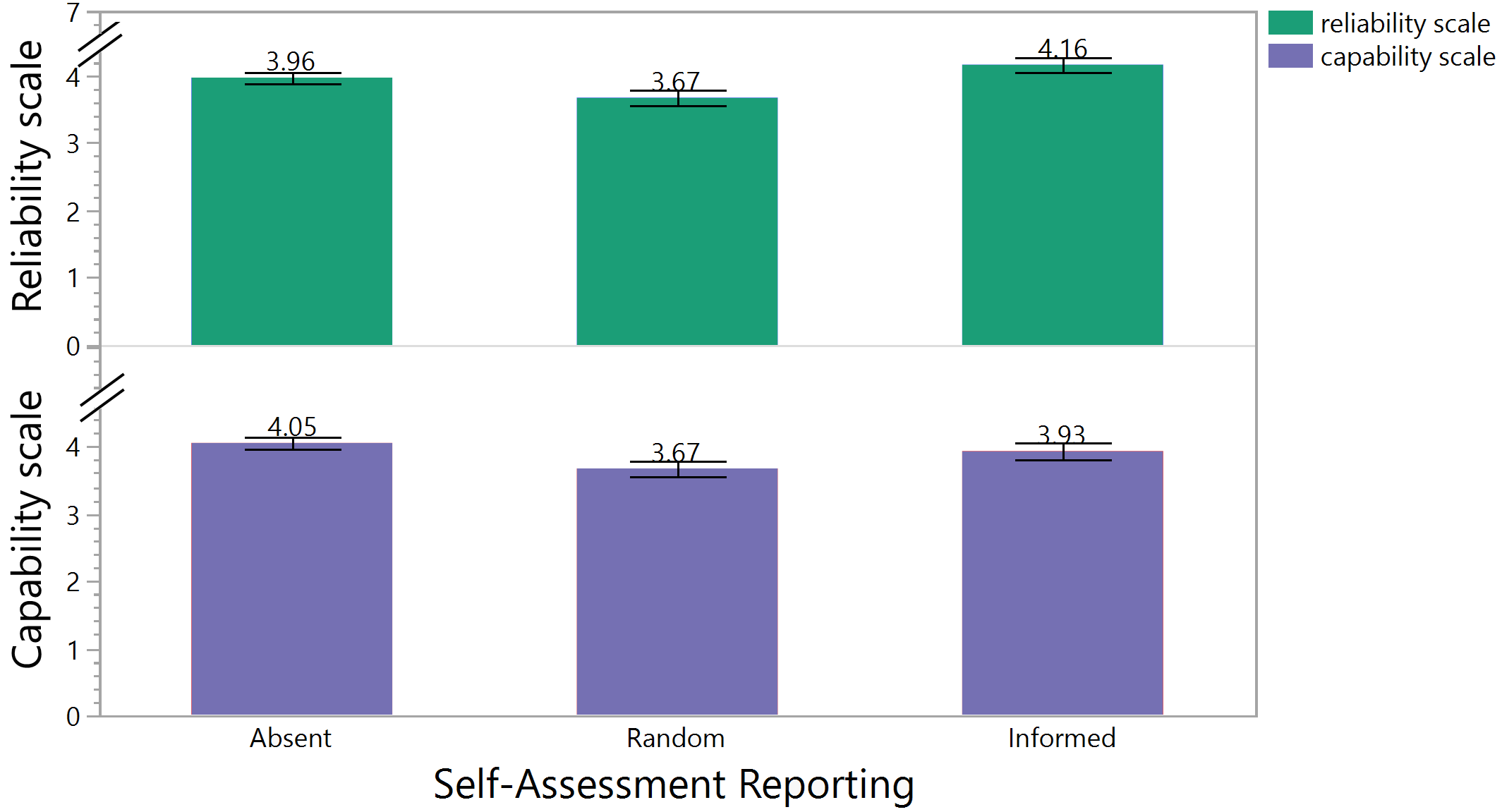}
    \caption{Bar plot of self-reported MDMT Reliability and Capability trust ratings for absent, random, and informed self-assessment reports showing a small improvement in reliability and capability trust between the informed and absent condition.}
    \label{fig:TRUST_VS_ACCURACY}
\end{figure}

\emph{RQ4: Does trust increase when the robot presents an \textbf{informed report}?} Fig. \ref{fig:TRUST_VS_ACCURACY} shows a bar plot of self-reported MDMT Reliability and Capability trust ratings for absent, random, and informed self-assessments reported to the participant. We conducted a one-tailed independent t-test to test whether the informed reporting affected participants self-reported MDMT capability and reliability trust scores and found that in both cases trust increased but not significantly for capability, $t(598.98)=1.55$, $p=0.0605$, and reliability, $t(598.98)=0.66$, $p=.2539$. For reference the figure includes random self-assessment reports, however we did not include it in our analysis for this specific question.

We believe that the small differences in trust were partly due to the trust scale used. The MDMT Performance dimension, which is composed of Reliability and Capability scales, measures trust related specifically to robot performance. We made the assumption that the participants would consider performance to include both the robots physical performance (navigating the grid world) and the robots reasoning performance (self-assessment reporting). This however may not be the case, and a future research direction for us is to investigate trust metrics and scales aimed specifically at an agents cognitive reasoning ability.

\emph{RQ5: Does the presence of robot self-assessment reports cue the participant to change autonomy to the appropriate level?} We broke this question into two parts:

\begin{figure}[htbp]
    \centering
    \includegraphics[width=0.4\textwidth]{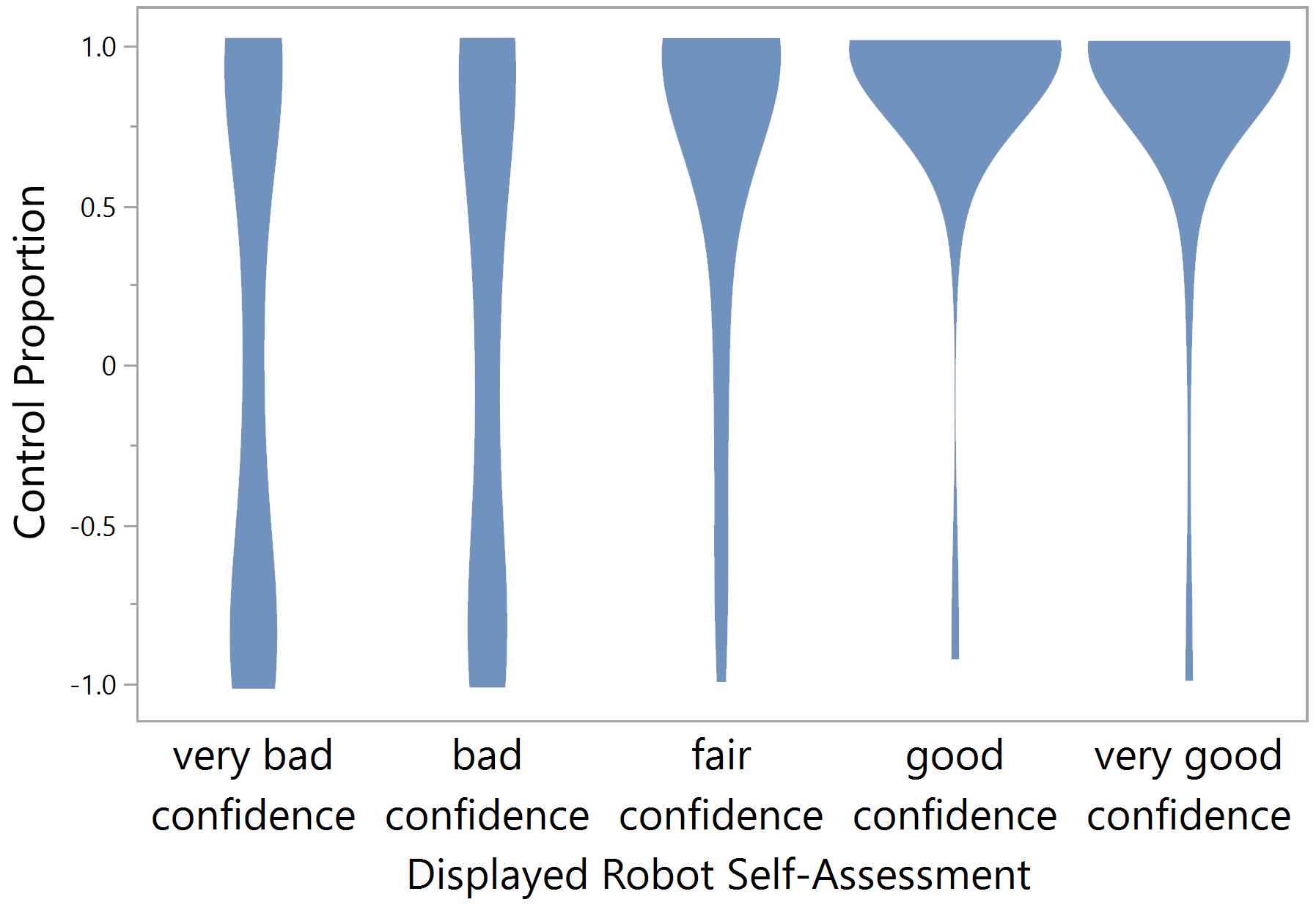}
    \caption{Box plot of control proportion for all robot self-assessment reports showing alignment between robot reports and control proportion. Positive control proportion indicated more robot control while negative control proportion indicated more participant control.}
    \label{fig:CONT_PROP_VS_ALL_SINGLE_REPORTS}
\end{figure}

\emph{All self-assessment reports}: We conducted a one-way analysis of variance (ANOVA) to test whether the presence of a robot self-assessment report affected the control proportion and found significant effect of report presence on control proportion, $F(4, 52.57)$, $p <.0001$. Post-hoc comparisons using Tukey’s HSD test revealed difference of control proportion between tasks reported as very bad and [fair, good, very good] $(p< .0001)$ and tasks reported as bad  and [fair, good, very good] $(p<.0001)$. This indicated that the team was more likely to change to a level of autonomy appropriate for the displayed robot self-assessment report. However here we also saw a tendency for the participant to follow the report regardless of its accuracy, which underscores the importance of a robot being able to accurately self-assess. This data is shown in Fig. \ref{fig:CONT_PROP_VS_ALL_SINGLE_REPORTS}.

\begin{figure}[htbp]
    \centering
    \includegraphics[width=0.4\textwidth]{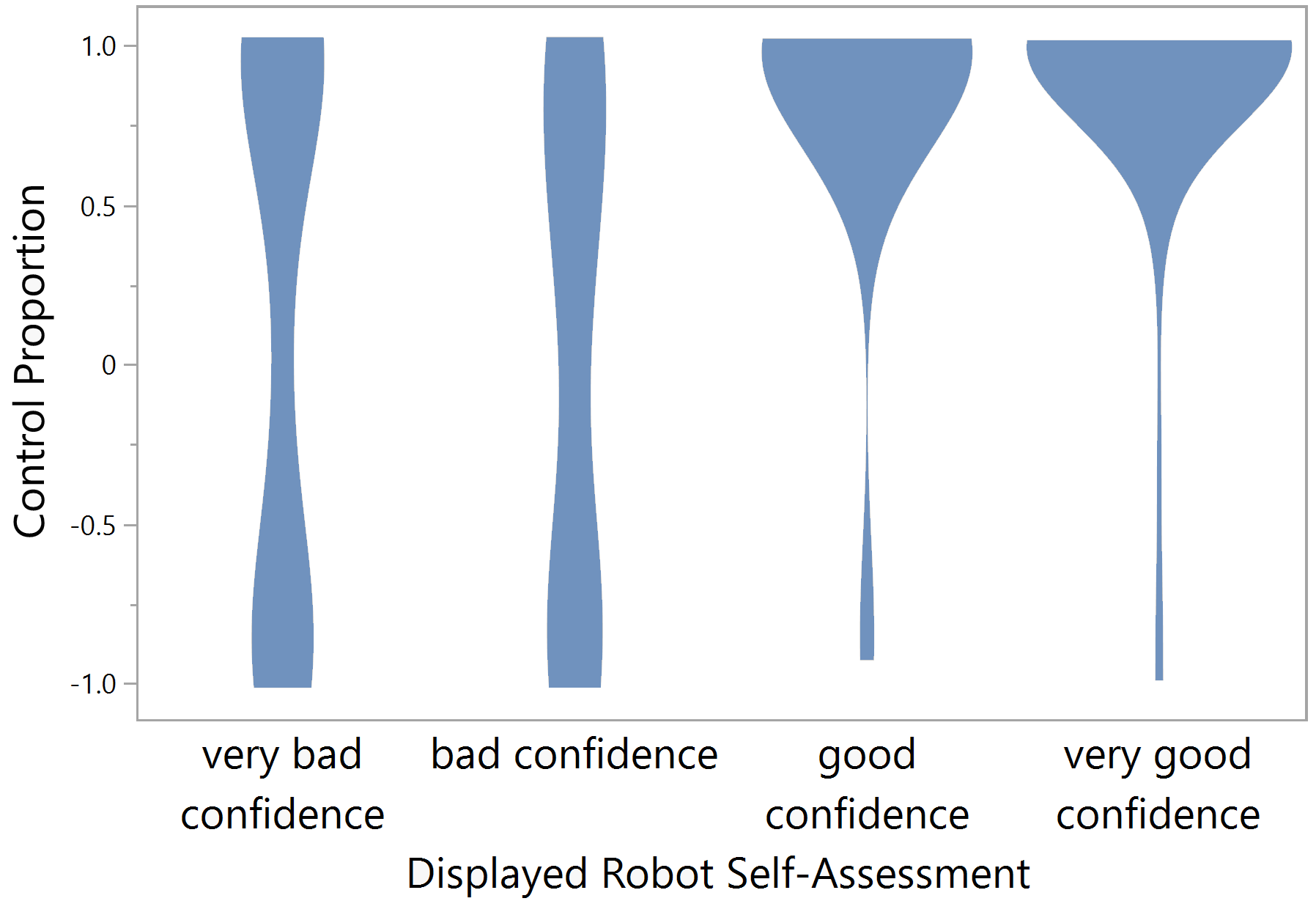}
    \caption{Box plot of control proportion for all informed robot self-assessment reports showing alignment between the robots reported confidence and control proportion. Positive control proportion indicated more robot control while negative control proportion indicated more participant control.}
    \label{fig:CONT_PROP_VS_ACCURATE_SINGLE_REPORTS}
\end{figure}

\emph{Informed self-assessment reports}: We conducted a one-way analysis of variance (ANOVA) to test whether the presence of informed robot self-assessment reports affected the control proportion and found significant effect of informed reports on control proportion, $F(3, 29.04)$, $p <.0001$. Post-hoc comparisons using Tukey’s HSD test revealed difference of control proportion between tasks reported as very bad and [good, very good] $(p< .0001)$ and tasks reported as bad and [good, very good] $(p<.0001)$. This data is shown in Fig. \ref{fig:CONT_PROP_VS_ACCURATE_SINGLE_REPORTS}. Similar to the previous case, this indicated that participant were more likely to change to a level of autonomy appropriate for the displayed robot self-assessment report. However control proportion also shows us that Automatic Control mode was still used (though much less frequently) for tasks with poor self-assessments (very bad, bad). Further exploration should investigate possible biases towards using the automation in the presence of robot self-assessments.

\section{FUTURE WORK} \label{FUTURE_WORK}
We have shown the value of robot proficiency self-assessments to human-robot teams given simple short duration tasks. However, during more complex tasks or longer interactions an \textit{a priori} self-assessment can quickly become stale, leading to the human inadvertently push a system beyond its competency bounds. Additionally, control proportion as described here assumed a fixed proficiency, and as such may be inappropriate for interactions in which the robots proficiency changes over time. In order capture the dynamic nature of realistic team interactions, our future work will focus on methods for real-time robot self-assessments and the development of performance metrics suited for online and longer term interactions.

\addtolength{\textheight}{-9cm}

\section{CONCLUSIONS}
Reporting of a robot's \textit{a priori} proficiency self-assessment can provide valuable information to improve a human teammate's decision making process. In this study we found that the reporting of robot self-assessment, in the form of rewards based FaMSeC Outcome Assessment, resulted in a decrease in task failures and small improvements to the participants self-reported trust. An \textit{a priori} robot self-assessment essentially aligns the participants task expectations with the robots belief in its abilities. Our objective trust metric, control proportion, gave us insight into the participants choice of autonomy allocation: tasks in which the robot reports lower self-confidence in meeting user expectations showed an increase in participant control, while tasks in which the robot reported higher self-confidence in meeting user expectations showed an increase in autonomous control. This work is an incremental step in improving trust and performance within a human-robot team.

\section{ Acknowledgments}
The authors would like to thank Dr. Brett Israelsen (Raytheon Technologies Research Center) for his help reviewing this document, and for the many informative discussions relating to trust, machine self-confidence, and human subjects studies.

\bibliography{ref}
\bibliographystyle{aaai}

\end{document}